# Noise Effects in Fuzzy Modeling Systems:Three Case Studies

P.J. COSTA BRANCO and J.A. DENTE
Laboratório de Mecatrónica
Instituto Superior Técnico
Av. Rovisco Pais 1049-011 Lisboa Codex
PORTUGAL
pbranco@alfa.ist.utl.pt    http:\\pbranco.ist.utl.pt

*Abstract:* - Noise is source of ambiguity for fuzzy systems. Although being an important aspect, the effects of noise in fuzzy modeling have been little investigated. This paper presents a set of tests using three well-known fuzzy modeling algorithms. These evaluate perturbations in the extracted rule-bases caused by noise polluting the learning data, and the corresponding deformations in each learned functional relation. We present results to show: 1) how these fuzzy modeling systems deal with noise; 2) how the established fuzzy model structure influences noise sensitivity of each algorithm; and 3) whose characteristics of the learning algorithms are relevant to noise attenuation.

## 1  Introduction

One permanent problem in systems' modeling, despite the use of emergent methodologies as fuzzy logic, neural networks, or genetic algorithms, is the presence of perturbations on acquired learning data, noise. It always appears caused by measurement errors or by other sources perturbing the signals. Hence, noise becomes an important factor establishing the quality of data, being ambiguity source to the learning mechanisms [12]. When designing fuzzy models and choosing between a diversity of learning algorithms, aspects as computational time, learning performance, model simplicity, and mainly low noise sensitivity, dictate our choice of which will be a well-performed fuzzy system to operate in real environments. In references [7-8], we started to investigate the two first aspects, computational time and learning performance, when automatic modeling two experimental drive systems using three well-known fuzzy modeling methods: the simplified algorithm proposed by Wang and Mendel in [5], a cluster-based algorithm motivated by [4], and the neuro-fuzzy algorithm presented in [6]. The third aspect, model simplicity, is related with the structure considered to the fuzzy model. In [9], we proposed a new clustering technique, the Autonomous Mountain Method, and its application in establish a more suitable structure to the fuzzy model. However, these previous studies demonstrated that, when handling with real and complex environments as with drive systems, it is difficult to separate noise effects from other potential causes when occurs a poor fuzzy approximation. To separate and better understand these effects, this work uses a simple linear function, a plane, and its modeling employing the same three fuzzy algorithms to study the noise effects on fuzzy models.

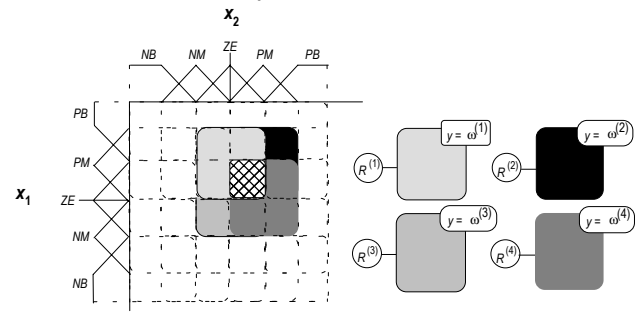

**Fig.1 Domain partition by the simplified algorithm.**

## 2  The Fuzzy Modeling Algorithms

This section overview the main features of each algorithm which are relevant to indicate how noise can modify their learning performance.

### 2.1  The simplified fuzzy algorithm

This algorithm was proposed by Wang and Mendel in [5]. Using uniformly distributed and symmetric triangular membership functions, the input-output space is subdivided into a rule set. Fig. 1 illustrates this domain partition for a bi-dimensional example where each operating region is covered by four rules.

The simplified algorithm uses the *maximum* operator to select the degree to which two fuzzy sets match. Thus, only the examples that significantly match the antecedent part of each rule are used to infer its conclusion value. Fig. 2 shows this process where it is selected, for a certain variable $x_i$, the fuzzy set linguistically representing the numerical value $x_i'$. Two membership degrees $\mu_{NM}(x'_i)$ and $\mu_{ZE}(x'_i)$ are attributed from fuzzy sets *NM* and *ZE*, respectively. The choice of the highest degree by *max*() operator indicates the fuzzy set *NM* as that one better characterizing the numerical value $x'_i$.

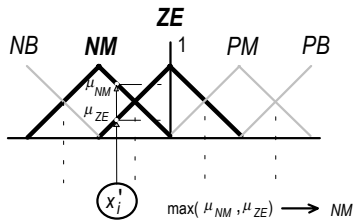

**Fig.2 Best set characterizing the numerical value $x'_i$.**

The *max*() operator limits the number of examples employed by the algorithm to infer each rule conclusion. The examples used will be those ones situated in the domain region where their membership values, in the fuzzy set being considered to represent its linguistic term, are superior to a 0.5 degree. Fig. 3(a) shows this restriction when extracting the conclusion of the rule defined by fuzzy sets *PM* and *NM*. The simplified algorithm will use only the examples marked in black since all have a membership degree higher than 0.5 in fuzzy sets *PM* and *NM*.

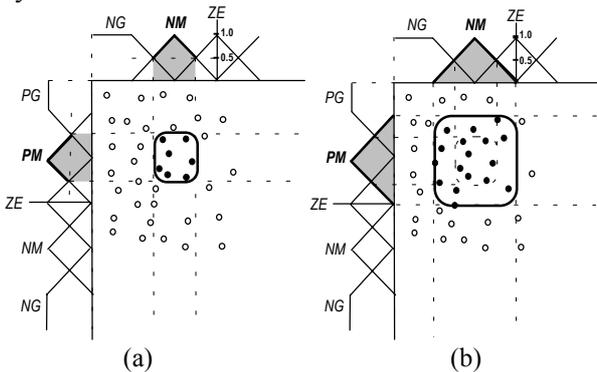

**Fig.3 (a) Examples used by the simplified algorithm. (b) Examples used considering the cluster concept.**

The simplified algorithm computes each conclusion based only in the information supplied by the best example that has the highest implication degree among all examples contained in the rule region. Moreover, if the best example is perturbed by noise, the imprecision in one example, producing an implication degree near 1.0, will induce an incorrect value as conclusion, independent from the other learning examples also be valid. To attenuate these drawbacks, the cluster idea can be introduced in the simplified algorithm. This idea, illustrated in Fig. 3(b), allows the algorithm to employ all examples situated in the considered domain region, also introducing a weighting mechanism that, actuating on the examples, reduces the distortion caused by noise in the fuzzy model extracted.

## 2.2 The cluster-based algorithm

Improving one more step about the simplified fuzzy algorithm, the cluster-based one uses the concept where each learning example contributes to more than one rule with a certain membership degree. The membership functions, initially settled in a symmetric triangular type, are uniformly distributed as previously. With a clustering idea, each example contributes with a different degree to the computation of each rule conclusion value. This allows the fuzzy algorithm to employ a higher number of examples only limited by the width of each triangular function. Fig. 3(a) shows the examples considered before and after using the cluster idea in Fig. 3(b).

Fig. 4 shows the algorithm's evolution concerning its capability to use the available examples in the learning set. Figure shows three initial clusters symbolized by rectangles filled with different colors: white, grey and black. Each rectangle delimits a certain domain region from which the algorithm acquires the examples used to extract the conclusion of the rule related with the referred cluster. Fig. 4 also shows the regions corresponding to the *max*() operator, having white examples, and the domain regions corresponding to the cluster idea with black examples.

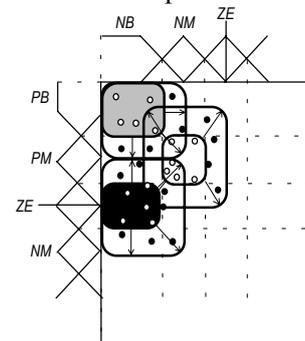

**Fig. 4. Simplified algorithm evolution concerning their capability of using the available learning examples.**

The cluster-based algorithm employs an average mechanism on the black examples which are weighted by their membership degrees at the considered cluster. The average mechanism allows a better noise attenuation, reducing possible distortions in the learned fuzzy model as it is shown in the next sections. Different from the simplified one, the cluster-based algorithm can use more examples to extract each rule conclusion. This also increases its learning performance and allows the use of smaller learning sets. References [1] and [4] introduce the algorithm details and show some developments of it.

## 2.3 The neuro-fuzzy algorithm

The neuro-fuzzy algorithm consisted in that one introduced by Wang in [3] and [6]. It uses gaussian membership functions covering all universe of discourse and divides it in equal parts. This algorithm, different when it were used membership functions of triangular type, can employ all available learning examples to extract each conclusion, and not only those ones restricted in the region covered by triangular membership functions as previously shown in Fig. 4. The neuro-fuzzy algorithm can also include a first stage where the previous algorithm initializes the conclusions. The conclusions are then tuned by a gradient descent method like the back-propagation technique to minimize the error between the estimated output of the fuzzy model and the measured one. Two parameters are necessary to be fixed: the learning parameter α and the number of learning iterations $K$. The three fuzzy modeling algorithms were initially used by the authors in the study of automatic modeling electromechanical systems. Results of our investigations, reported in [8] and [11], pointed out their potentialities as well their drawbacks. These investigations motivated the study presented in this paper to understand why and how the algorithms are affected by noise polluting the learning data.

## 3 Functional Relation

To better visualize and separate the effects of noise in each algorithm, we consider the modelization of a simple functional relation as the plane $z = x + y$ displayed in Fig. 5(a). The learning examples were supposed to have been acquired during a first training test and are shown in Fig. 5(b). This set of examples is initially considered not polluted by noise. It is also important note that the examples are not equally distributed through the domain, being concentrated in some regions and more sparse in other ones.

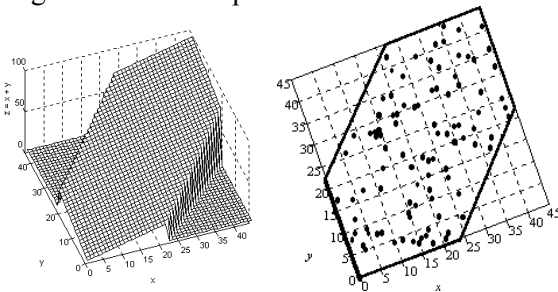

**Fig.5 (a) Relation $z = x + y$. (b) Data without noise.**

In our analysis, noise signal is simulated adding to each plane variable $x$, $y$, and $z$, a random value between ±10% of each their nominal value, which is an acceptable practical percentage of noise. Fig. 6 shows the plane function when noise pollutes the data. From this plane, a second learning set presenting examples perturbed with noise is collected.

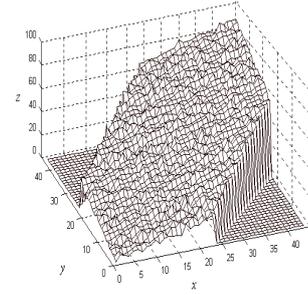

**Fig. 6. Functional relation with noise perturbation.**

To examine noise influence in the algorithms, results are displayed showing the difference signal between the plane fuzzy model obtained from the "clean" learning set and the fuzzy model extracted from the "noisy" learning set. Therefore, the effects of noise distorting the extracted fuzzy rules will be better exposed.

## 3 Verification of Noise Effects in the Fuzzy Simplified Algorithm

Figs. 7(a) to 7(d) show the effects of noise when using the fuzzy simplified algorithm. Results were obtained increasing the number of membership functions of the fuzzy model from only 3 fuzzy sets to $x$, $y$ and 13 fuzzy sets to $z$ in Fig. 7(a), to a number of 9 fuzzy sets partitioning $x$ and $y$ variables, continuing to attribute 13 fuzzy sets to the output variable $z$ in Fig. 7(d). Note that the number of 13 functions of the output variable was maintained constant during the tests to focalize our analysis of noise influence at function's domain level. This reduces its influence on the output variable, which allowed us to separate both effects. Results in Fig. 7(a), obtained with three domain partitions, show that with the simplest fuzzy model structure, noise does not significativelly affected the modeling process. Its performance was mainly imposed by the reduced complexity of the fuzzy model, since by attributing only 3 fuzzy sets to $x$ and $y$ variables the model acquires very general rules. Increasing of domain partition, the rules become more specific. However, as shown by the results in Figs. 7(b) to 7(d), higher is the model complexity, higher are the noise effects deforming the fuzzy model. Fig. 8 shows in more detail the results using the fuzzy model with nine partitions. Fig. 8(a) shows the rules extracted without noise perturbation. Fig. 8(b) shows the corresponding learned function. The presence of noise in learning data rearranges the examples to other domain regions. This concentrates more examples in certain domain areas and open large empty areas in other ones. To verify the consequence of these effects in the rule-base, the distribution of the rules in Fig 8(a), obtained without noise, are compared with the distribution extracted from the noisy learning set shown in Fig 9(a), where some new and incorrect rules were obtained. The distortions caused on the fuzzy model can be visualized comparing the relation obtained without noise in Fig. 8(b) with that one extracted with noise and shown in Fig. 9(b). When noise perturbs the conclusion values, it introduces conflict interpretations in the rule-base resulting then in high modeling errors.

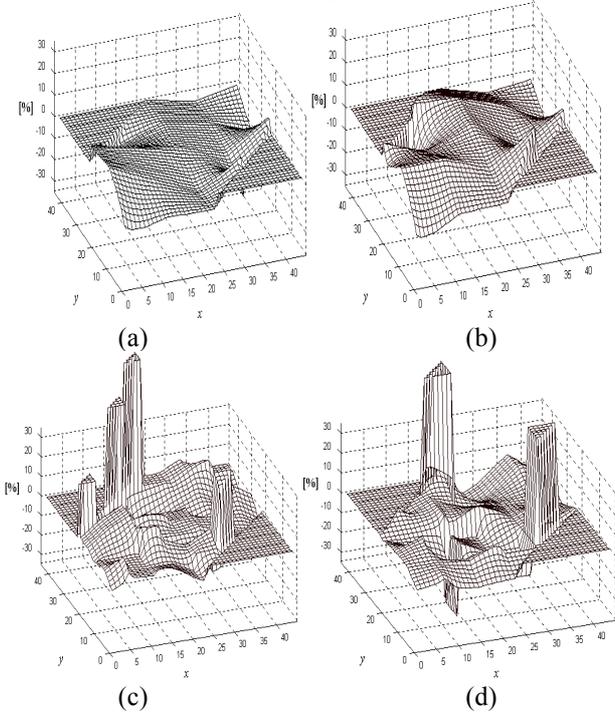

**Fig.7 Simplified algorithm. Difference between "clean" and noisy models (a) $3\times 3$, (b) $5\times 5$, (c) $7\times 7$ and (d) $9\times 9$ model.**

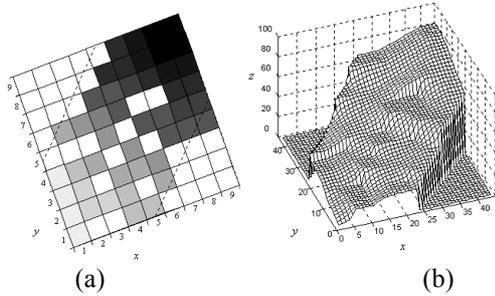

(a)          (b)

**Fig.8 Simplified algorithm *without* noise perturbation and the $9\times 9$ model. (a) Extracted rules. (b) Learned relation.**

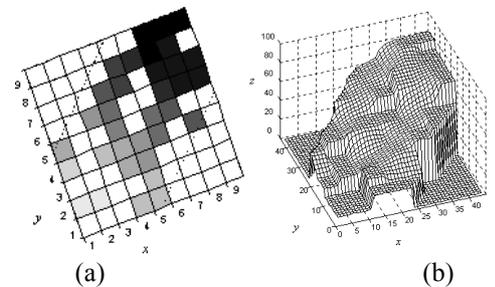

(a)          (b)

**Fig.9 Simplified algorithm *with* noise and the $9\times 9$ model (a) Extracted rules. (b) Learned relation.**

The simplified algorithm has a local feature. Its performance depends of how the domain is covered by the examples, and also the number of examples used to extract each conclusion [2, 10]. If the total amount of examples is high and noise magnitude is limited, this will not significantly affect the examples' distribution by the domain, which thus attenuates the introduction of false rules. Fig. 10 shows this aspect using again the model with nine partitions. The result shows the difference between the relation extracted from a large and noisy learning set, with 400 examples, and the relation obtained from a clean set with also 400 examples, demonstrating the low noise influence. Noise actuates in two directions. First direction is related with the examples' distribution. Second direction is related with the distortion made by noise on the output variable *z*. This indicates the necessity of to have some ponderation mechanism actuating on the examples to attenuate the distortions in the output variable.

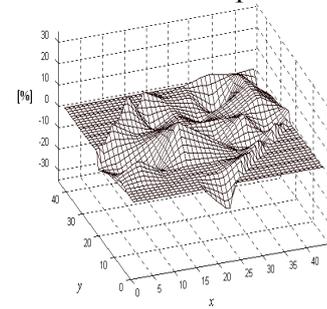

**Fig. 10. Results with the $9\times 9$ model with a larger and "noisy" data set.**

## 3 Verification of Noise Effects in the Cluster-Based Algorithm

### 3.1 Triangular membership functions

The cluster-based algorithm uses a larger number of examples to extract each rule conclusion, also employing a weighted mechanism to its computation. To better verify its noise attenuation capacity, the noise signal is increased to ±30% of each variable's nominal value, instead of ±10%, as shown in Fig. 11.

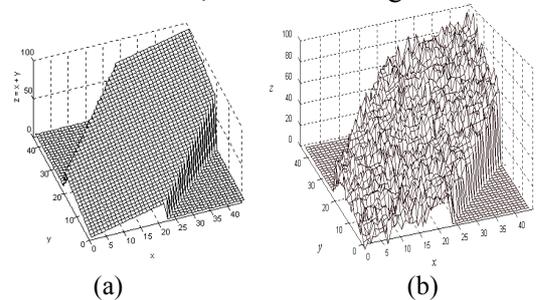

(a)          (b)

**Fig. 11. (a) Plane without noise. (b) Plane perturbed with noise in a ±30% level.**

As in the previous section, we verify the noise attenuation as the complexity of the fuzzy model structure is increased. Figs. 12(a) to 12(d) show the results obtained with the fuzzy models using three, five, seven and nine partitions. Despite the use of a higher noise magnitude, the cluster-based algorithm shows better noise attenuation than that one obtained with the simplified algorithm. This is verified comparing Figs. 7(a) to 7(d) with the results in Figs. 12(a) to 12(d). Also note that the noise influence continues to increase with the fuzzy model complexity from three to nine fuzzy partitions.

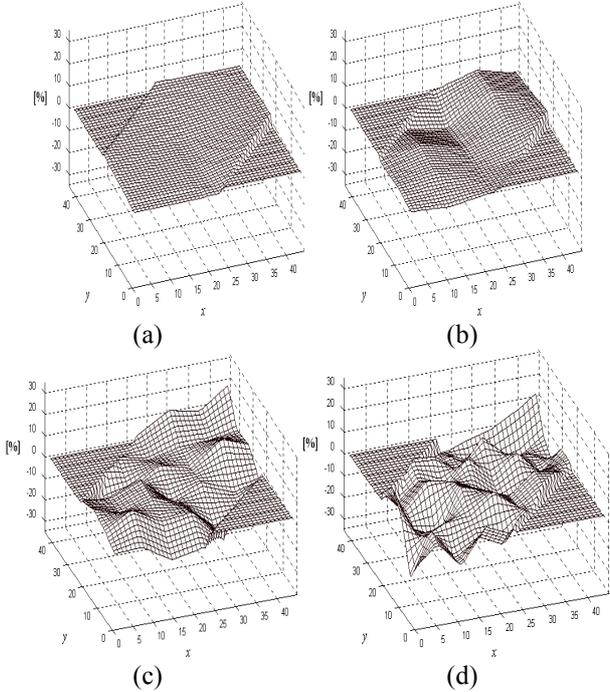

**Fig.12 Difference between "clean" and "noisy" models (a)** $3\times 3$**, (b)** $5\times 5$**, (c)** $7\times 7$ **and (d)** $9\times 9$ **model.**

### 3.2 Gaussian membership functions

This section presents an intermediate stage before using the neuro-fuzzy algorithm. It consists in to substitute the triangular functions in the cluster-based algorithm by gaussian ones, as shown in Fig. 13. The algorithm uses the same increasing complexity for the fuzzy model structure, with the gaussian functions symmetrically distributed by the universe of discourse of each input variable.

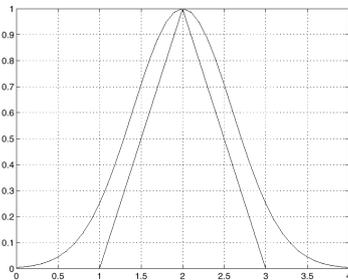

**Fig.13 Triangular functions to gaussian ones.**

When using gaussians, noise attenuation capacity of the cluster-based algorithm increases as shown in Figs. 14(a) to (d). It uses all examples that are available to compute each conclusion, instead of only use a small subset of them. The distortions in the conclusion values caused when noise redistributes the examples to other areas in the domain are also decreased. When modeling with triangular functions, one has only a local use of the learning examples, while gaussians allow a global use of all available data, permitting better noise attenuation.

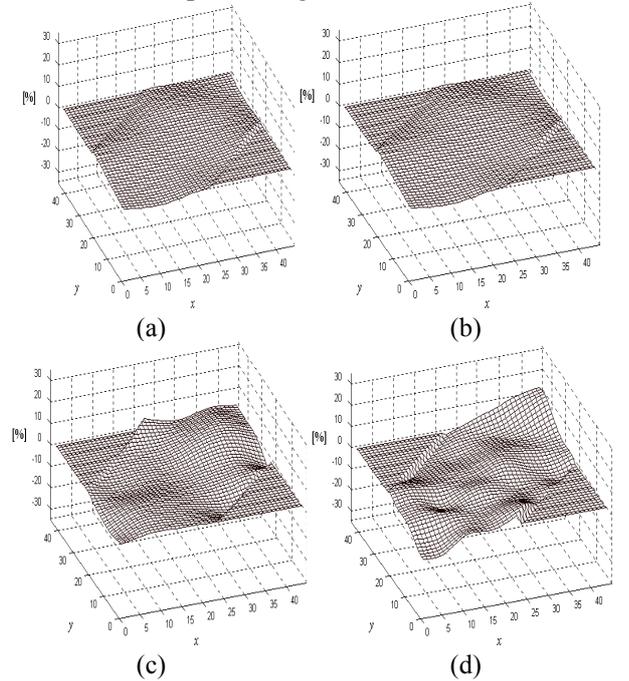

**Fig. 14. Error signal between the "clean" and "noisy" models (a)** $3\times 3$**, (b)** $5\times 5$**, (c)** $7\times 7$ **and (d)** $9\times 9$ **model.**

## 4 Verification of Noise Effects in the Neuro-Fuzzy Algorithm

The neuro-fuzzy algorithm follows the same model structure of the cluster-based one using gaussian functions, and uses a supervised learning methodology. As shown in Figs. 15(a) to 15(d), the neuro-fuzzy algorithm presents the highest noise attenuation of all since it shows also better interpolation results. Despite its highest performance, the use of high values to the learning rate parameter makes the fuzzy model begins to "learn" example by example and so to present overfitting, which incorporates noise in the extracted conclusions. To visualize this effect, a test with a higher learning rate value than 0.1, used in the results of Fig. 15, was performed. Fig. 16(a) shows the results when using a learning rate value of 0.8. Fig. 16(b) displays the extracted relation for a more critical learning rate of 0.95, revealing the increased deforming effects induced by noise perturbation in the learning process. The learning rate value is then decreased in the manner that the algorithm can attenuate noise effects through its interpolation capability. Fig. 16(c) shows these results to a test where the learning rate value was decreased to its original low value of 0.1, severely decreasing the fuzzy model distortion.

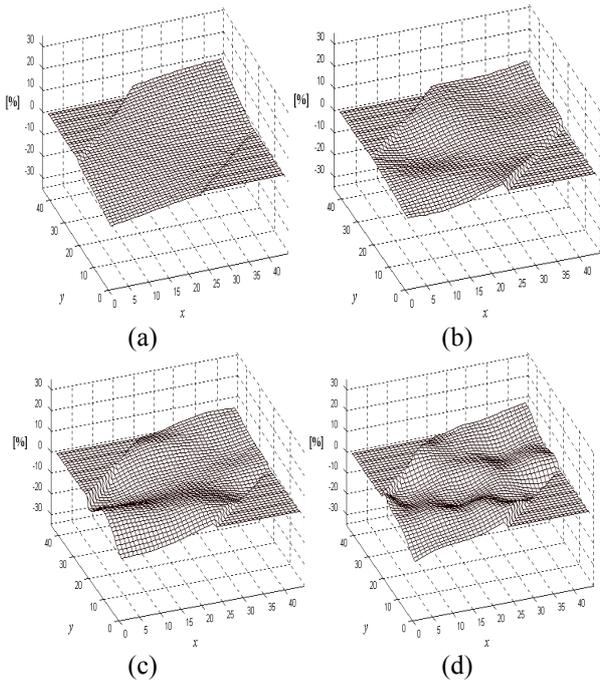

**Fig.15** Error between the "clean" and "noisy" fuzzy models **(a)** $3\times 3$, **(b)** $5\times 5$, **(c)** $7\times 7$ **and (d)** $9\times 9$ **model.**

## 4 Conclusion

This paper focused on how and why fuzzy modeling systems are affected when learning data is polluted with noise. We evaluated its effects studing three well-known fuzzy modeling algorithms applied to a simple plane function. This procedure allowed us to illustrate and separate the causes increasing, or decreasing, noise sensitivity of each algorithm. Results showed that the established fuzzy model structure and certain characteristics of the learning algorithm are directly related with its noise attenuation capacity. The algorithms were analyzed by their evolution in noise attenuation, indicating the neuro-fuzzy algorithm as that one most robust to noise. The analysis was motivated by our experimental work in [8]. Although preliminary, it allowed to reveal how and why each algorithm deal with noise, helping us in the design of a good fuzzy model to operate in real-time fuzzy control systems.

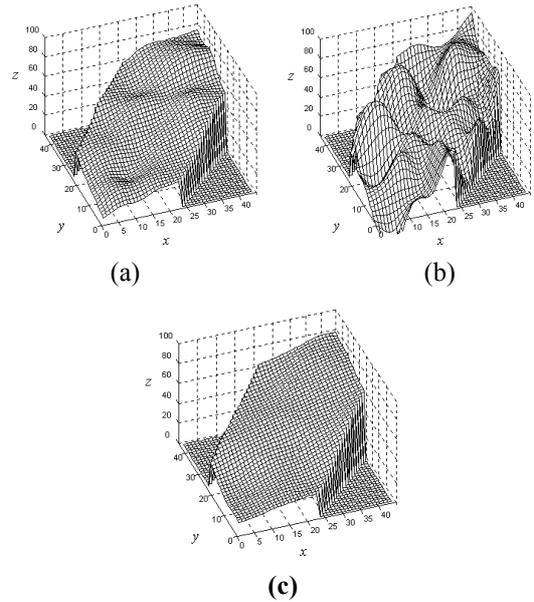

**Fig.16** Relations from the neuro-fuzzy algorithm with noise. Learning rate values **(a) 0.8, (b) 0.95 and (c) 0.1.**